\documentclass[conference]{IEEEtran}
\IEEEoverridecommandlockouts
\usepackage[utf8]{inputenc}
\usepackage[T1]{fontenc}
\usepackage{microtype}
\usepackage{cite}
\usepackage{amsmath,amssymb,amsfonts}
\usepackage{booktabs}
\usepackage{array}
\usepackage{graphicx}
\usepackage{textcomp}
\usepackage{xcolor}
\definecolor{green}{rgb}{0,0.45,0}
\usepackage{algorithm}
\usepackage{algpseudocode}
\newif\ifincludelifecycle
\includelifecycletrue
\usepackage{tikz}
\usetikzlibrary{arrows.meta,positioning,calc,fit}
\usepackage[draft,bookmarks=false]{hyperref}
\usepackage{url}
\let\citep\cite
\newcommand{\pkg}{\texttt{agent.pkg}}
\newcommand{\profile}{\texttt{workflow.v0}}
\newcommand{\sys}{WebAgents}
\title{WebAgents: Unbundling Agent Creation, Hosting,\\and Execution for an Agent-Native Web
\thanks{This work was supported by EU Horizon Europe through SmartCHANGE (Grant 101080965) and by the Swiss National Science Foundation through XAI-PAC (Grant PZ00P2\_216405).}
}
\author{
\IEEEauthorblockN{
Dario Fenoglio\IEEEauthorrefmark{1}\IEEEauthorrefmark{2},
Martin Gjoreski\IEEEauthorrefmark{1}, 
Marc Langheinrich\IEEEauthorrefmark{1}, 
Pradyumna Chari\IEEEauthorrefmark{2},
and Ramesh Raskar\IEEEauthorrefmark{2}
}
\IEEEauthorblockA{
\IEEEauthorrefmark{1}Università della Svizzera italiana (USI), Switzerland\\
\IEEEauthorrefmark{2}Massachusetts Institute of Technology (MIT), United States\\
Corresponding author: \texttt{dario.fenoglio@usi.ch}
}
}
\hypersetup{pdftitle={WebAgents: Unbundling Agent Creation, Hosting, and Execution for an Agent-Native Web},pdfsubject={Workshop-style research report; RQ1 checkpoint 23eac90; three-package RQ2 checkpoint 9d78ca5},pdfkeywords={Agentic Web, portable AI agents, OpenClaw, NANDA, agent packages}}
\begin{document}
\maketitle

\begin{abstract}
Large language models are moving from isolated assistants toward autonomous agents that can plan, use tools, and coordinate with others, motivating an emerging Agentic Web of independently developed agents and services.
Realizing this vision requires agent behavior to move across creators, hosts, and runtimes without coupling publication to a particular execution environment or transferring authority over user resources. We introduce \sys{}, which unbundles these roles through a signed \pkg{} that carries portable workflows, skills, schemas, and resource requirements. 
An OpenClaw-based agent-native browser/runtime 
verifies the package, obtains authorization, and binds it at run time to user-controlled models, tools, permissions, and budgets, while hosts distribute the public program and may retain private services server-side.
We demonstrate this contract with three unrelated workflows executed unchanged under local and API-backed models, including literature review with live services and a website-published travel agent using semantic tools and user-selected personal resources. 
In a controlled suite, \sys{} explicitly blocks all 716 injected contract violations while accepting all 295 matched benign controls, with median automated pre-execution processing below 0.5 s for all three packages.
We conclude by outlining a path toward NANDA-based discovery and interoperable package ecosystems while retaining execution authority in the user's runtime.
\end{abstract}


\section{Introduction}
Large Language Models (LLMs) are increasingly moving beyond passive content generation toward autonomous agents that can plan, invoke tools, maintain state, and coordinate with others. These capabilities are already being explored across domains ranging from scientific discovery to enterprise automation, suggesting a shift from isolated AI applications toward ecosystems of interacting agents~\citep{webofagents,react,autogen}. As these systems mature, the Web can become a substrate not only for humans accessing pages and services, but also for agents discovering capabilities, delegating tasks, and acting across organizational boundaries.
We refer to this emerging setting as the \emph{Agentic Web}: an open environment where independently developed agents, tools, and services can be discovered, composed, and executed~\citep{yangAgenticWebWeaving2025}. Realizing this vision, however, requires more than increasingly capable models. It requires infrastructure that allows agent behavior to move across creators, hosts, and runtimes without coupling publication to a particular execution environment or transferring authority over user resources.

Consider a researcher sharing a literature-review workflow. Hosting it as a service makes the publisher responsible for orchestration and inference, while distributing a script leaves each recipient to configure models, credentials, and tools. Neither approach, by itself, provides a common boundary between the published workflow and the recipient's authority.

We argue that coupling agent creation, hosting, and runtime limits an open agent ecosystem. We therefore separate three roles: (i) a creator produces a portable agent program; (ii) a host makes that program reachable and may retain private services server-side; and (iii) a user-controlled runtime verifies and executes it using authorized resources. 
This separation lets creators distribute expertise without operating every invocation, while users retain control over the models, credentials, and budgets used on their behalf. With a common execution contract, creation tools, hosting providers, and runtimes can evolve independently.

However, separating these roles is not sufficient. A downloaded agent may request access to private data, invoke remote tools, or consume a substantial inference budget. The package must therefore describe behavior without granting itself the authority to perform it; the receiving runtime decides which requests to accept and enforces those decisions throughout execution. In our framing, an \emph{agent-native browser} executes a publisher-defined program rather than reconstructing behavior from human-facing pages. The goal is a web designed for agents, not only agents designed to navigate the existing web~\citep{dengMind2WebGeneralistAgent2023,zhouWebArenaRealisticWeb2023}. Website automation can remain one of its tools, but is not its defining function.

To explore this separation, we introduce \sys{}, a portable execution framework built around a signed agent package and an OpenClaw-based browser/runtime. \sys{} realizes the three roles around a shared \pkg{} artifact: creators produce and sign portable packages, a host distributes them independently of execution, and the OpenClaw agent-native browser/runtime binds them at run time to the user's authorized models, tools, permissions, and budgets. This connects signed behavior, user authorization, and concrete model/tool bindings in an auditable execution contract. We demonstrate the design with three workflows executed unchanged under local and API-backed model bindings, and use controlled experiments to test whether the contract rejects violations and measure its automated pre-execution cost. We then identify requirements for portable execution and outline a path toward discovery and richer execution profiles. 

With this work, we hope to contribute toward a more open and interoperable Agentic Web in which reusable agent behavior can move across providers and runtimes while execution authority remains with the user.

\section{Related work}
\label{sec:related}
\paragraph{Agent interoperability and discovery}
The Web of Agents identifies messaging, interaction, state management, and discovery as foundations for cross-ecosystem collaboration~\citep{webofagents}; Agent-OSI develops communication and settlement interfaces for a decentralized agent ecosystem~\citep{xuAgentOSIInteroperabilityArchitecture2026}. MCP standardizes access to tools and context, including consent requirements~\citep{mcp}, while A2A supports task exchange between independently implemented agents~\citep{a2a}. Work on agent infrastructure motivates identity and discovery beyond fixed service endpoints~\citep{upgrade}; NANDA separates a lean index from verifiable AgentFacts and endpoint metadata~\citep{beyond}. These mechanisms let agents find and invoke capabilities. Our focus is the additional step of transferring an agent program for execution under the recipient's authority.

\paragraph{Portable agent behavior}
Agent Skills packages reusable instructions, scripts, and resources that agents can load across compatible environments~\citep{skills}, while Paper2Agent exposes research methods and workflows as reusable MCP tools~\citep{miaoReimaginingResearchPapers2026}. Open Agent Specification defines a framework-independent representation of agents, workflows, data flow, and tool interfaces, with adapters that execute the same specification across multiple agent frameworks~\citep{openagentspec}. \sys{} moves portability to a different boundary: portable execution across independently operated creators, hosts, and runtimes. 
A signed program is distributed independently of execution, while the receiving runtime selects and authorizes the concrete models, tools, permissions, and budgets used on its behalf. Thus, portability does not require adopting the publisher's framework, infrastructure, or execution authority.

\paragraph{Package identity and provenance}
JAR combines manifests, content digests, and signatures~\citep{jar}, while in-toto verifies signed evidence of software production steps~\citep{torresIntotoProvidingFarmtotable2019}. The Agent Packaging Standard also specifies agent manifests, distribution, and optional provenance verification~\citep{aps}. These approaches establish artifact identity and provenance. \sys{} extends this boundary into execution: the verified package is bound to the user's approved execution configuration, and the same binding must still hold when execution begins. 

\paragraph{Runtime control and enforcement}
D'Agents combined authenticated programs with resource policies and restricted execution~\citep{dagents}. For LLM agents, AIOS separates applications from managed models, tools, and access control~\citep{meiAIOSLLMAgent2025}. AgentSpec enforces event--condition--action rules at runtime~\citep{wangAgentSpecCustomizableRuntime2026}, and CaMeL uses explicit control and data flow with capability policies against prompt injection~\citep{camel}. These systems show the importance of mediating 
execution when external capabilities are involved.
\sys{} brings publication, authorization, and runtime enforcement into one verifiable lifecycle, connecting this 
control to a portable, independently published agent program. We evaluate this lifecycle through unchanged-package executions, contract-violation tests, and pre-execution timing.

\section{Unbundling through a portable execution contract}
\label{sec:contract}
An open Agentic Web needs publication and execution to scale independently. Reusing an agent should not require the creator to operate its inference or the user to adopt a publisher-specific resource environment. We therefore define unbundling as \emph{independent substitutability}: creators, hosting providers, and compatible runtimes can change without rewriting the public agent program. A provider can offer several roles; the shared contract keeps the others replaceable.

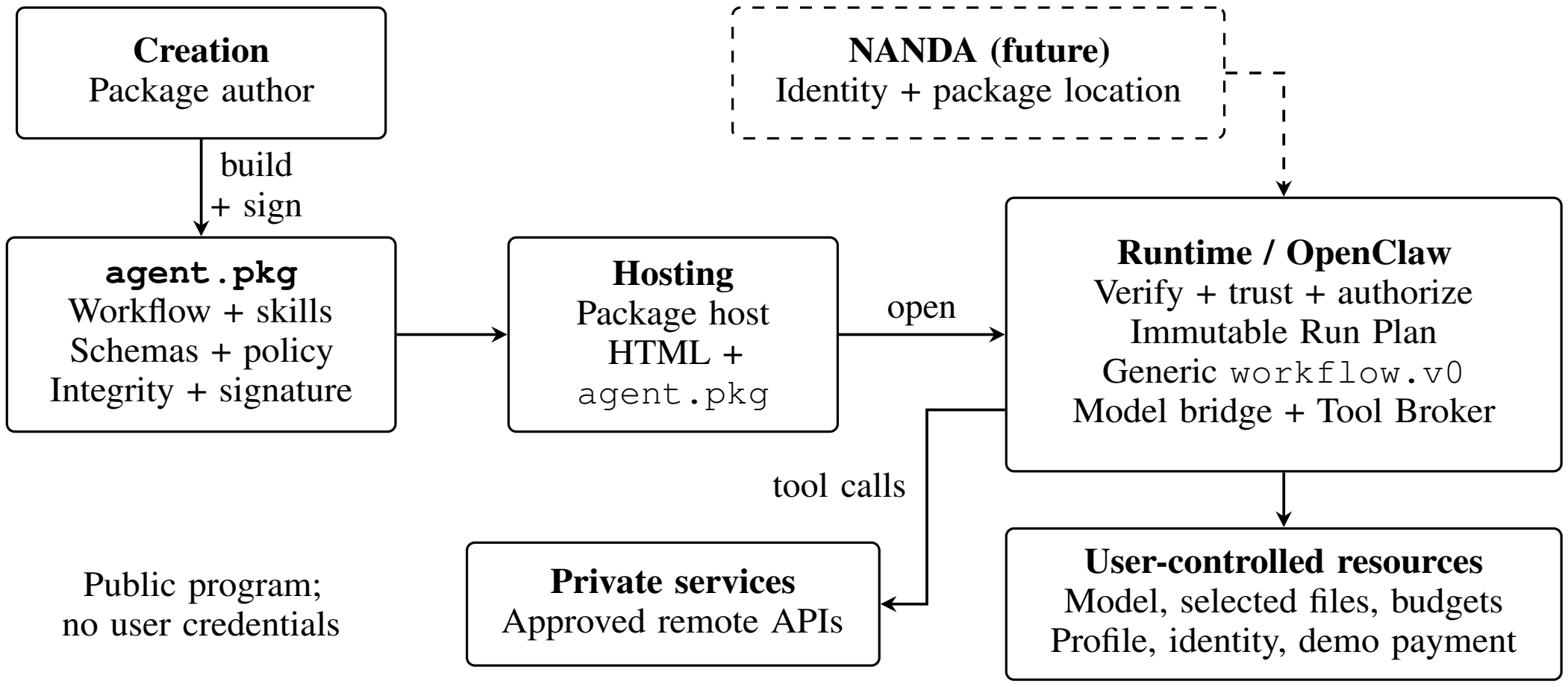
\begin{figure*}[t]
  \centering
  \begin{tikzpicture}[x=1pt,y=1pt,
    box/.style={draw,rounded corners=2pt,align=center,font=\small,inner sep=5pt,line width=.6pt},
    arr/.style={-{Stealth[length=4pt,width=4pt]},line width=.7pt},
    lab/.style={font=\small,align=center,inner sep=2pt}]
  \node[box,text width=83pt,minimum height=33pt] (author) at (48,142) {\textbf{Creation}\\Package author};
  \node[box,text width=88pt,minimum height=49pt] (package) at (48,76) {\textbf{\texttt{agent.pkg}}\\Workflow + skills\\Schemas + policy\\Integrity + signature};
  \node[box,text width=73pt,minimum height=49pt] (host) at (167,76) {\textbf{Hosting}\\Package host\\HTML + \pkg{}};
  \node[box,dashed,text width=114pt,minimum height=33pt] (nanda) at (244,142) {\textbf{NANDA (future)}\\Identity + package location};
  \node[box,text width=130pt,minimum height=69pt] (runtime) at (321,76) {\textbf{Runtime / OpenClaw}\\Verify + trust + authorize\\Immutable Run Plan\\Generic \texttt{workflow.v0}\\Model bridge + Tool Broker};
  \node[box,text width=94pt,minimum height=31pt] (services) at (167,8) {\textbf{Private services}\\Approved remote APIs};
  \node[box,text width=130pt,minimum height=31pt] (resources) at (321,8) {\textbf{User-controlled resources}\\Model, selected files, budgets\\Profile, identity, demo payment};
  \draw[arr] (author) -- node[lab,right]{build\\+ sign} (package);
  \draw[arr] (package) -- (host);
  \draw[arr] (host) -- node[lab,above]{open} (runtime);
  \draw[arr,dashed] (nanda.east) -| (runtime.north);
  \draw[arr] (runtime.south) -- (resources.north);
  \draw[arr] ([yshift=-19pt]runtime.west) -- (231,57) -- (231,8) -- (services.east);
  \node[lab,anchor=east] at (228,38) {tool calls};
  \node[lab,text width=92pt] at (48,8) {Public program;\\no user credentials};
  \end{tikzpicture}
  \vspace{-1pt}
  \caption{Separation of publication and execution in \sys. The package transports behavior and requirements; it does not transport user authority. Solid paths include experimental website-advertised discovery. The dashed discovery path is a proposed NANDA integration: its output would enter the same verification and authorization process as a direct package URL.}
  \label{fig:architecture}
  \vspace{-5pt}
\end{figure*}

\paragraph{Creation, hosting, and runtime}
The creator specifies the workflow, skills, and requirements in a signed \pkg{}. The host makes the package reachable and may separately expose private services. The browser/runtime inspects the package and supplies the user’s chosen model, approved tools, and execution limits (Figure~\ref{fig:architecture}). This separates the cost of making an agent available from the cost of running it: inference is provisioned by the runtime, while private data and business logic can remain behind service interfaces. Users can reuse a configured environment across independently authored agents, and creators can improve workflows without operating each user's execution stack. As the ecosystem grows, shared interfaces would let creation tools, hosting services, and runtimes improve independently rather than requiring one platform to provide all three.

\paragraph{Requirements are not permissions}
A literature-review package may require structured model outputs, sufficient context capacity, and scholarly search. These requirements describe what execution needs; the runtime determines which concrete resources satisfy them and obtains permission to use them. The resulting contract binds the signed program to those approved resources and limits for a particular run. This is the connection between portability and control: the program can be shared across environments, while credentials and authority remain with each receiving runtime. Compatibility is explicit, and different model bindings can produce different answers.

\section{The WebAgents architecture}
\label{sec:architecture}
The prototype implements the separation in Figure\hspace{-1pt} \ref{fig:architecture}\hspace{-1pt} through a data-only package, a hosting layer, and a user-controlled OpenClaw runtime containing the browser control plane and mediated execution engine. The host distributes the package and may expose private services, while the runtime verifies, authorizes, binds, and executes it. Hosting a package does not grant access to runtime credentials or tools.

\subsection{A signed program, not a capability grant}
\label{sec:package}
An \pkg{} is a deterministic archive containing a manifest, workflow graph, markdown skills, input/output schemas, tool declarations, and policies. The manifest identifies the package version and execution profile, and declares the model capabilities and resources needed to run the workflow. The current format contains data interpreted by \profile{}; it does not load package-supplied Python, shell scripts, or native modules.

Package verification uses SHA-256 content hashes and an Ed25519 publisher signature over the canonical integrity statement. The browser also records a hash of the exact archive bytes. Parsing is bounded and rejects ambiguous or unsafe structures, including duplicate archive names, path traversal, duplicate authoritative JSON keys, and remote schema loading. Inspection does not execute the workflow. This matters because authorization must be based on a stable description of what will run, not on behavior already triggered while opening it.

A valid signature binds content to a signing key; it does not establish that the signer is trusted or that the workflow is safe. In the prototype, the runtime independently pins accepted signer keys under an explicit scope. The package is inspectable by design, so secrets do not belong inside it. Provider credentials stay with the runtime, while proprietary host-side logic remains accessible only through approved service interfaces.

\subsection{From inspection to an immutable Run Plan}
\label{sec:plan}
The browser opens an explicit HTTPS package URL or an approved local artifact. When a user instead provides a website URL, the browser can also discover an \pkg{} published by that site. HTTPS acquisition is size-bounded and rejects redirects and private-network targets. It checks the package structure, integrity, signature, and supported runtime before evaluating publisher trust. The user then grants or denies permissions and chooses a compatible, routable model. Tool selection has a further boundary: a declared tool, an available runtime implementation, and an approved binding are distinct objects. A general permission to perform scholarly search does not authorize arbitrary network requests.

These decisions produce a \emph{Run Plan}: a hash-bound description of the exact package snapshot, signer and trust state, authorization, selected model, approved tool bindings, and personal-resource grants. A separate \emph{Run Invocation} binds task input to that plan. Immediately before execution, the runner re-verifies the snapshot and current authority. If the package, trust decisions, or bindings no longer match the plan, execution is refused rather than silently updated. This closes the gap between the artifact the user inspected and the artifact actually executed.

The plan is preparation, not an independent permission token. Execution still passes through the trusted runner and its checks. A distinct \emph{Run Record} captures completion status, usage, artifacts, and audit information without mutating the plan. Separating these objects makes it possible to compare the authorized configuration with what a particular run consumed and produced. \ifincludelifecycle{}Appendix~\ref{app:lifecycle} summarizes the complete package-to-result lifecycle.\fi

\subsection{Generic execution with runtime-controlled resources}
\label{sec:runtime}
The \profile{} interpreter executes an acyclic graph using six node types: \texttt{llm}, \texttt{tool}, \texttt{map}, \texttt{join}, \texttt{validate}, and \texttt{output}. Model and tool nodes use schema-checked interfaces; map nodes create bounded workers; join nodes combine results; validation nodes check declared structural and cross-record constraints. Skills and orchestration are package data, with no dispatch on package names. Workers receive copied state views, and validated writes are committed after successful execution waves. Failed waves do not partially commit shared workflow state.

Model calls pass through a model bridge to exactly the model recorded in the Run Plan. The package cannot access provider credentials or silently switch to a fallback. Abstract requirements are checked against runtime capability information, while returned structured data is independently validated.

Tool calls pass through the runtime Tool Broker. Each binding fixes an adapter and its contract; the package cannot choose arbitrary request URLs, headers, filesystem paths, or shell commands for execution. For the Literature Review package, the runtime provides adapters for Crossref search, a remotely hosted Citation Resolver, and user-selected PDF snapshots. The resolver returns scholarly metadata and BibTeX. 
For PDF access, the runtime exposes opaque file identifiers rather than local paths. WebAir adds four operator-configured HTTP bindings for its airline services. These are example runtime bindings, not built-in workflow operators: other packages can use different adapters without changing the interpreter.

Packages can also request user-controlled resources by type, such as a travel profile or passenger identity. The runtime exposes eligible resources, the user selects and authorizes one, and that choice is bound to the Run Plan. The package therefore expresses what it needs without selecting or carrying the user’s private resource itself.

The runtime enforces token and call budgets, worker concurrency, and deadlines outside the prompts. It reserves resources before dispatch and accounts for actual usage afterward. 
Cancellation propagates to model and tool providers where supported. These controls bound execution, but neither cancellation nor failed state commits imply rollback of an external service's completed side effects.

Execution\hspace{-1pt} emits a locally hash-linked audit trail with pro- venance and usage artifacts, recording run configuration and actions rather than certifying its answers. Enforcement assumes a trusted OpenClaw runtime and approved adapters. A malicious instruction can still misuse an approved capability, as agent-risk and prompt-injection evaluations illustrate~\citep{ruanIdentifyingRisksLM2023,debenedettiAgentDojoDynamicEnvironment2024}.

\section{Evaluation}
\label{sec:evidence}
Our evaluation separates three questions. The live case studies test whether unchanged packages execute under different model bindings. RQ1 tests whether the contract blocks violations without rejecting matched benign executions. RQ2 measures the local cost of its automated pre-execution checks.
\ifincludelifecycle{} Additional implementation lessons appear in Appendix~\ref{app:lessons}. \fi 
The \sys{} implementation, evaluated packages, and experimental artifacts are publicly available.\footnote{\url{https://github.com/dariofenoglio98/WebAgents}}

\subsection{Three packages, one interpreter}
Table~\ref{tab:runs} summarizes \emph{Meeting Agenda}, \emph{Literature Review}, and \emph{WebAir} executed through \profile{} under both a local Ollama model and an OpenAI API model. For each workflow, the package archive and task input are identical across the two runs, so changing the model binding does not require repackaging the agent. \emph{Meeting Agenda} plans a meeting, creates three parallel role-based perspectives, and synthesizes a structured agenda. Both 
runs produced agendas whose items sum to the requested 25 minutes. It uses no external tools, providing a case distinct from literature review without an application-specific runner.

\begin{table*}[t]
\caption{Acceptance runs through the same generic interpreter. Within each workflow, local and API runs use identical package bytes, task inputs, permissions, runtime limits, and tool bindings. 
Times are single-run workflow durations.}
\label{tab:runs}
\centering
\small
\setlength{\tabcolsep}{4pt}
\begin{tabular}{@{}lrrrrrr@{}}
\toprule
 & \multicolumn{2}{c}{\textbf{Meeting Agenda}}
 & \multicolumn{2}{c}{\textbf{Literature Review}}
 & \multicolumn{2}{c}{\textbf{WebAir}} \\
\cmidrule(lr){2-3}\cmidrule(lr){4-5}\cmidrule(l){6-7}
 & \textbf{Local} & \textbf{API} & \textbf{Local} & \textbf{API} & \textbf{Local} & \textbf{API} \\
\midrule
Model & Qwen3-30B & GPT-5.6 Luna & Qwen3-30B & GPT-5.6 Luna & Qwen3-30B & GPT-5.6 Luna \\
Model calls & 5 & 5 & 18 & 18 & 3 & 3 \\
External tool calls & 0 & 0 & 6 & 6 & 4 & 4 \\
Personal-resource reads & 0 & 0 & 0 & 0 & 1 & 1 \\
Peak parallel workers & 3 & 3 & 3 & 3 & 3 & 3 \\
Reported tokens & 11,346 & 5,198 & 100,352 & 68,398 & 11,577 & 7,716 \\
Workflow duration (s) & 154.3 & 29.4 & 1,489.8 & 417.6 & 81.7 & 18.6 \\
Audit events & 92 & 102 & 572 & 678 & 241 & 247 \\
Model cost (USD) & N/A & 0.0147 & N/A & 0.2005 & N/A & 0.0094 \\
\bottomrule
\end{tabular}
\end{table*}


\emph{Literature Review} adds live services to the same runtime. Its package defines search, screening, citation resolution, evidence extraction, synthesis, writing, and verification. Both runs address the same task on prompt-injection defenses with the same approved tool bindings. 
Each run made three Crossref searches and three Citation Resolver calls and produced a related-work draft with bibliographic, evidence, unresolved-item, and verification artifacts. 
All 18 model calls in each run used the model selected in its Run Plan.

\emph{WebAir} adds a website-native case: a fictional airline publishes a signed \pkg{} alongside its human-facing webpages. After discovering and verifying the package, the agent can interact with WebAir through its agent-native tools for flight search, fare rules, booking quotes, and simulated purchase, rather than reconstructing these actions from the human-facing site. The matched search runs each read the selected travel profile and make four live service calls: one flight search and three fare-rule requests. 

All six runs completed with valid audit chains.
Together, they show that the same packaged workflows can execute under local and API-backed model bindings without changing the package itself. This demonstrates portability across model-provider bindings within the same OpenClaw runtime. 

\subsection{RQ1: Contract enforcement}
\label{sec:enforcement}
Successful workflows do not establish that their execution boundaries are enforced. We therefore inject violations of the contract's declared guarantees across eight families (Table~\ref{tab:enforcement}). Each template specifies the expected refusal stage and an objective forbidden boundary before the run. Examples include a tampered package reaching a prepared plan, an unplanned model's answer being committed, or a call being charged below its honest usage.

\paragraph{Setup and metrics}
Across Meeting Agenda and Literature Review, spanning tool-free and tool-enabled workflows, the suite contains 716 distinct violations from 61 templates and 295 matched benign controls that follow the same execution paths without the violated condition, preventing trivial reject-all behavior.
Each case exercises the Agent Browser, control plane, and runtime, with deterministic stand-ins for external dependencies. We count a violation as detected only when the forbidden boundary is not reached and the contract explicitly refuses execution, while the controls measure whether benign executions remain accepted. To identify which mechanisms provide this protection, we repeat the suite under eight ablations, each disabling one enforcement component. \ifincludelifecycle Full threat-model, outcome, simulator, and ablation details are given in Appendix~\ref{app:enforcement}.\fi

\begin{table*}[t]
\caption{Contract enforcement: percentage of injected violations explicitly blocked before their forbidden boundary. Ablations are defined in the text; bold rates mark family-level drops. All 295 benign controls are accepted in every profile; incidental failures and harness errors are zero. The overall row is weighted by this catalog's case counts. 
}
\label{tab:enforcement}
\centering
\small
\setlength{\tabcolsep}{3.4pt}
{\begin{tabular}{@{}lrrrrrrrrrrr@{}}
\toprule
 & & & & \multicolumn{8}{c}{Removed enforcement component} \\
\cmidrule(lr){5-12}
Family & Viol. & Controls & Full & Verif. & Trust & Author. & Compat. & \shortstack{Plan\\reval.} & \shortstack{Runtime\\med.} & \shortstack{Output\\val.} & Resource \\
\midrule
Package verification & 191 & 134 & 100 & \textbf{48.7} & 100 & 100 & 100 & 100 & 100 & 100 & 100 \\
Publisher trust & 14 & 12 & 100 & 100 & \textbf{14.3} & 100 & 100 & 100 & 100 & 100 & 100 \\
Authorization & 34 & 12 & 100 & 100 & 100 & \textbf{47.1} & 100 & \textbf{52.9} & 100 & 100 & 100 \\
Requirements / compatibility & 85 & 17 & 100 & 100 & 100 & 100 & \textbf{76.5} & 100 & 100 & 100 & 100 \\
Plan binding & 55 & 21 & 100 & 100 & 100 & 100 & 100 & \textbf{30.9} & 100 & 100 & 100 \\
Runtime mediation & 80 & 15 & 100 & 100 & 100 & 100 & 100 & 100 & \textbf{30.0} & 100 & 100 \\
Output / state integrity & 196 & 59 & 100 & 100 & 100 & 100 & 100 & 100 & 100 & \textbf{17.9} & 100 \\
Resource governance & 61 & 25 & 100 & 100 & 100 & 100 & 100 & 100 & 100 & 100 & \textbf{18.0} \\
\midrule
\textbf{Overall} & \textbf{716} & \textbf{295} & \textbf{100} & 86.3 & 98.3 & 97.5 & 97.2 & 92.5 & 92.2 & 77.5 & 93.0 \\
\bottomrule
\end{tabular}}
\end{table*}

\paragraph{Results}
The full system explicitly blocked all 716 violations before their forbidden boundary, at the lifecycle stage specified in advance, and accepted all 295 controls. All 9,099 executions across the nine profiles were recorded without harness errors. Removing each component exposed violations in its corresponding family. For example, detection fell to 48.7\% without package verification, 30.9\% without plan revalidation, 30.0\% without runtime mediation, and 17.9\% without output validation. Every lost block became a silent violation, not an incidental failure.

The checks also overlap. Removing plan revalidation reduced authorization detection to 52.9\% because 16 authorization cases concern consent binding or expiry before the run deadline. Other layers continued to block subsets of the ablated families, including retained structural checks and checks at later boundaries. The matrix therefore identifies which cases depend on each component, rather than showing that one check can enforce the whole contract.

\subsection{RQ2: Pre-execution overhead}
\label{sec:overhead}
We measure the automated contract processing from signed package bytes already in memory to the interpreter's first-node dispatch boundary, before any node executes. To approximate the cost of local execution in the user-controlled runtime, we evaluate it on a local workstation: an Apple M3 Max (14 cores, 36 GB), with Python 3.13.5. Each of three runs includes 50 warmup iterations per package followed by 1,000 timed observations per package, giving 3,000 observations per package. Package order is randomized within each iteration. 
\ifincludelifecycle Appendix~\ref{app:overhead} gives the protocol and artifact sizes.\fi

\begin{table}[t]
\caption{Automated pre-execution processing (ms): pooled median (95th percentile), 3,000 observations per package.}
\label{tab:overhead}
\centering
\small
\setlength{\tabcolsep}{3pt}
\begin{tabular}{@{}lrrr@{}}
\toprule
Operation & \shortstack{Meeting\\Agenda} & \shortstack{Literature\\Review} & WebAir \\
\midrule
Package verification & 19.3 (20.4) & 152.7 (160.1) & 161.7 (168.4) \\
Authorization / binding & 1.7 (1.8) & 2.7 (2.9) & 4.0 (4.2) \\
Run Plan preparation & 2.1 (2.3) & 3.9 (4.2) & 6.5 (6.9) \\
Pre-execution revalidation & 35.8 (38.1) & 289.4 (301.1) & 316.1 (327.3) \\
\midrule
\textbf{Total} & \textbf{59.0 (62.5)} & \textbf{449.9 (465.0)} & \textbf{488.7 (504.7)} \\
\bottomrule
\end{tabular}
\end{table}

Median processing took 59.0\,ms for Meeting Agenda, 449.9\,ms for Literature Review, and 488.7\,ms for WebAir (Table~\ref{tab:overhead}), 
corresponding to less than 0.2\% of end-to-end workflow duration for Meeting Agenda and Literature Review, and 0.6--2.6\% for WebAir. The coefficient of variation of run medians was at most 0.41\%. Repeated verification and payload validation at the control plane, runner, and interpreter account for approximately 90\%, 96\%, and 96\% of the measured time.
The higher processing times for Literature Review and WebAir relative to Meeting Agenda are consistent with their larger archives, workflow graphs, and sets of declared tools. A separate component microbenchmark identifies schema and cross-reference validation as the largest contributor to package-open time; Ed25519 verification takes at most 0.39\,ms.


These measurements isolate the contract's local processing, not end-to-end startup latency. They exclude package download, human authorization time, model and tool calls, loopback HTTP transport, and runner process startup. 
Together, the experiments show rejection of the injected violations and sub-second automated preparation on this machine.

\section{Toward an open Agentic Web}
\label{sec:future}
The current prototype opens packages from explicit HTTPS URLs, experimental website-advertised links, or approved local artifacts. NANDA integration would let users discover candidate packages by identity and capability~\citep{beyond}, then enter the same verification and authorization pipeline. This extends how programs are found while preserving the execution contract.

\paragraph{Discovery without delegated execution authority}
This integration needs an explicit mapping from agent identity and AgentFacts to a package URL, version, and content digest. AgentFacts describes an agent; \pkg{} contains the program to execute. Neither a resolver result nor a signed capability assertion grants execution authority. The browser would still verify the downloaded artifact, evaluate trusted issuers, and obtain local consent. A certificate authority could bind publisher identities to signing keys, while evaluators attest to tested package properties~\citep{vc}. Runtimes could accept these credentials under their own issuer policies, checking scope and validity rather than manually approving every key. Freshness, revocation, and conflicting registry information need defined failure behavior. Federated discovery could then connect public and organizational registries without making one index the universal runtime authority.

\paragraph{An ecosystem needs conformance, not just more packages}
Open collaborative agent ecosystems require interoperability across messaging, interaction, state management, and discovery to avoid fragmentation across frameworks~\citep{webofagents,xuAgentOSIInteroperabilityArchitecture2026}. \sys{} extends this interoperability to the execution boundary: independently published agent programs should move across creators, hosts, and runtimes while execution authority remains with the receiving runtime.
Achieving this requires versioned schemas, stable execution-profile semantics, and 
conformance tests. Evaluation should identify the package digest, runtime configuration, and tested workload rather than attaching an unconditional quality label to an agent. Future work should execute unchanged packages across independently implemented runtimes and multiple model/tool configurations, measuring completion, resource use, and authorization violations. This would test substitutability across runtimes, extending the current model-binding and contract evaluations.


\paragraph{Broader execution and service integration}
Conversational and persistent-service profiles require explicit state lifetimes, renewed consent, cancellation, and budgets. Executable modules need an additional isolation boundary beyond the current 
data-only format. MCP tools, A2A delegation, and legacy-web automation could enter through 
runtime adapters~\citep{mcp,a2a}. 
A changed binding must be evaluated and authorized again, never substituted within an existing Run Plan. Broader data-flow controls inspired by CaMeL~\citep{camel} and confirmations for consequential actions could further constrain the use of approved capabilities. Scoped persistent memory, external attestations, and real payment integration with monetary spending controls are additional extensions.

\section{Conclusion}
We introduced \sys{} as an implementation of agent creation, hosting, and runtime separation. Its signed package carries the public program, while the OpenClaw-based runtime supplies and controls the authority to execute it. Three documented workflows show that the same packaged behavior can execute with both local and API-backed models, including an origin-published website agent using live services and runtime-held personal resources. In the controlled suite, \sys{} blocks all 716 injected violations and accepts all 295 benign controls, while the three-package benchmark measures median pre-execution costs of 59.0, 449.9, and 488.7\,ms. Together, these results support the feasibility of the execution contract and its enforcement within the evaluated scope.
The next challenge is to turn that working boundary into a shared contract across creators, hosts, runtimes, and discovery systems without treating portability as permission. More broadly, \sys{} points toward an open and interoperable Agentic Web in which reusable agent behavior can move across providers and runtimes while execution authority remains with the user.

\section*{AI Disclosure}
ChatGPT and Claude were used to assist with language editing, improving manuscript clarity, and coding support. All technical content, experiments, analyses, and references were verified by the authors.

\bibliographystyle{IEEEtran}
\bibliography{references}

\ifincludelifecycle
\appendices
\section{Package-to-result lifecycle}
\label{app:lifecycle}
Algorithm~\ref{alg:lifecycle} summarizes the contract in Section~\ref{sec:architecture}, including website-advertised acquisition. In the current prototype, a website can advertise an associated \pkg{} through an experimental same-origin \texttt{rel="agent-package"} link. NANDA discovery and issuer credentials remain future extensions. The lifecycle keeps artifact identity, resource authorization, and execution records distinct.

\begin{algorithm}[t]
\small
\caption{From package acquisition to an auditable result}
\label{alg:lifecycle}
\begin{minipage}{\linewidth}
\begin{algorithmic}[1]
\Require Package or website location, task input, and runtime policy; optionally a future NANDA identity.
\Ensure Preparation error or Run Record with status and available artifacts; existing plans unchanged.
\State \textbf{Discover.} Read an experimental same-origin website package link; \emph{future:} resolve identity through NANDA to AgentFacts and a package location/version/digest. Treat either result as metadata, not authority.
\State \textbf{Acquire.} Snapshot bounded package bytes from HTTPS or an approved local artifact without executing the workflow.
\State \textbf{Verify.} Check container, hashes, signature, schemas, and execution profile. Reject invalid packages.
\State \textbf{Evaluate trust.} Require a scoped signer pin; future issuer credentials can inform trust without replacing verification.
\State \textbf{Authorize and bind.} Obtain permissions, tool approvals, and personal-resource selections. Choose a compatible, routable model; refuse unavailable required capabilities.
\State \textbf{Prepare.} Freeze package, signer/trust, receipt, model, tool bindings, and resource identifiers/digests in a Run Plan. Bind input in a Run Invocation.
\State \textbf{Revalidate.} Lease the snapshot; recheck identity, authority, bindings, resource grants, model availability, and limits. Refuse stale plans.
\State \textbf{Execute.} Run the signed graph with copied state and approved bridges, enforcing permissions, schemas, budgets, concurrency, and deadlines.
\State \textbf{Validate and commit.} Validate outputs and declared relationships; commit writes only after successful waves. Record progress and usage.
\State \textbf{Handle interruption.} Stop affected work and propagate cancellation where supported, without rolling back completed external side effects.
\State \textbf{Finalize.} Emit Run Record, audit, usage, provenance, and available artifacts. Publish successful outputs after checks; preserve the Run Plan.
\end{algorithmic}
\end{minipage}
\end{algorithm}



\section{Controlled evaluation protocol}
\label{app:enforcement}
\subsection{Violation construction and outcome rules}
RQ1 (experiment 1.0.0, threat model 1.0, commit \texttt{23eac90}) uses Meeting Agenda 0.2.2 and Literature Review 0.2.5, not WebAir. Of 61 templates, 54 are enumerated and seven parametric. Seed \texttt{20261002} varies content edits, added members, context sizes, token-budget fractions, and schemas. Nine schema-mutation operators invalidate calibrated benign answers; duplicate parameters are rejected. Thirty-nine templates apply to both packages and 22 only to Literature Review, yielding 253 and 463 violations.

The eight families cover package tampering; untrusted/mis-scoped signers; missing, denied, replayed, or expiring consent; incompatible requirements or tool contracts; post-plan changes; unapproved capability use; malformed outputs or broken relations; and resource overruns or invalid usage. Each template fixes its assumption, refusal component/stage/codes, forbidden boundary, matched control, and ablation before the run. Controls restore the missing condition, use valid perturbations, or set limits at measured need; deduplication leaves 295 controls.

A boundary crossing is a silent violation even if an error follows. Otherwise, detection requires an explicit production refusal; unexplained failure is incidental. Refusal codes are checked against production source. Simulated-runtime refusals receive no credit; failed injections and harness errors are separate. No violation was detected only after its forbidden boundary.

\subsection{Execution and ablation scope}
The compiled plugin and \texttt{agentpkg} run from source with simulators for models, services, DNS, and PDF parsing. Workers refuse and audit real network, process, and canary-file access. The suite is offline and credential-free, with in-process control plane and runner; the runner uses the real bridge protocol. It tests contract logic, not OS isolation, live services, compromised runtimes, release rollback, side channels, availability, or independent-runtime interoperability.

Ablations use isolated source copies, exact-match Python/TypeScript patches, and control-plane policy switches. Workers verify their profile; no production option exposes ablations. Verification retains parsing/schema prerequisites, runtime mediation retains static workflow checks, and output validation retains JSON parsing/lookups. Unremoved sites also block four root-level schema mutations and one tool-timeout variant. These gaps limit ablation coverage: remaining detection does not imply protection after all checks of that type are removed.

All 716 refusals matched the predeclared stage; component attribution matched 710 and refusal codes 704. Six runtime-version cases are labeled package validation; six trust-withdrawal cases reach control-plane preflight before the publisher check. Boundary outcomes are unchanged. Development exposed under-charging of malformed usage reports, fixed before this run. Expectations were corrected and three templates split before freezing the catalog, without changing outcome rules. Smoke, development, and pre-fix runs are excluded.

Rates describe this finite catalog: variants share templates and the family mix determines the pooled rate. Controls rule out reject-all behavior here, not false rejections on arbitrary benign workloads.

\subsection{Pre-execution timing protocol}
\label{app:overhead}
RQ2 (benchmark 1.1.0, commit \texttt{9d78ca5}) uses the artifacts in Table~\ref{tab:packages}. Only the new three-package series is pooled. A long-lived control plane uses \texttt{time.perf\_counter\_ns} with garbage collection enabled. Per run, 50 warmups precede five batches of 200 iterations per package, with package order randomized using seed \texttt{20261001}. Three runs yield 9,000 observations and 450 warmups, with none of the timed observations dropped. Code, configuration, fixtures, package hashes, and machine match across runs.

Benchmark fixtures describe 11 tool bindings and six personal resources. For WebAir, all seven bindings are approved but only the travel profile is granted; identity, payment, and purchase permissions remain denied. Grant/denial propagation is checked at the runner. Literature Review uses 0.2.4 in Table~\ref{tab:runs} and 0.2.5 in RQ1/RQ2.

\begin{table}[t]
\caption{Signed artifacts in RQ2. 
}
\label{tab:packages}
\centering
\small
\setlength{\tabcolsep}{4pt}
\begin{tabular}{@{}lrrr@{}}
\toprule
Property & \shortstack{Meeting\\Agenda} & \shortstack{Literature\\Review} & WebAir \\
\midrule
Version & 0.2.2 & 0.2.5 & 0.3.0 \\
Archive bytes & 21,580 & 224,302 & 167,193 \\
Payload files & 17 & 66 & 58 \\
Graph nodes & 7 & 25 & 34 \\
Declared tools (remote / local) & 0 & 3 (2 / 1) & 7 (4 / 3) \\
Permissions (personal-resource) & 2 (0) & 5 (0) & 7 (3) \\
\bottomrule
\end{tabular}
\end{table}

Eight timestamps delimit package verification, authorization, model/tool compatibility, plan preparation, runtime revalidation/lease, runner re-verification, and workflow start. Table~\ref{tab:overhead} groups stages 2--3 as authorization/binding and 5--7 as pre-execution revalidation. Groups and totals are directly timed; medians need not sum. Preconditions are checked outside timers. Guards reject model/tool calls, network access, and process creation. Human decisions are pre-recorded; file reads are untimed.

Pooled p95 uses linear interpolation. Run-median coefficients of variation are 0.14\%, 0.31\%, and 0.41\% for the three totals and at most 0.94\% per stage. Runs began only when one-minute load was at most 3.0; none was discarded and no inference ran concurrently. Component measurements attribute most package-open time to schema/cross-reference validation; Ed25519 takes at most 0.39\,ms. Absolute timings also depend on runtime configuration and machine/session conditions.

\section{Implementation lessons}
\label{app:lessons}
\paragraph{Resource assumptions are part of the contract}
Earlier package versions passed deterministic fixtures but failed on the local model backend. Conservative token reservations exceeded declared budgets, and local reasoning exceeded deadlines chosen around hosted inference. Search breadth also exceeded output limits. The corrections changed bounded workload, timing, budgets, and skill instructions while preserving runtime checks. A transferable graph is therefore insufficient: a package also needs resource requirements that a runtime can assess. This makes incompatibility visible instead of hiding it behind a model substitution or relaxed limits.

\paragraph{Output contracts must cover composition and failure}
An earlier review run reused local evidence identifiers across workers, causing the join to discard three genuine records. The current runs use paper-scoped evidence identifiers, avoiding this collision. An earlier query returned insufficient grounded evidence; validation rejected unsupported prose instead of producing a valid no-results artifact. These cases show why output contracts must cover cross-worker relationships and empty-result behavior, not only individual schemas. Such checks cannot establish review completeness or whether claims follow from the evidence.

\fi
\end{document}